\documentclass[letterpaper, 10 pt, conference]{ieeeconf}  

\IEEEoverridecommandlockouts                              
\usepackage{graphics} 
\usepackage{epsfig} 
\usepackage{mathptmx} 
\usepackage{times} 
\usepackage{amsmath} 
\usepackage{amssymb}  
\usepackage{hyperref}
\usepackage{caption}

\begin{document}
\title{\LARGE \bf
An Auto-tuning Framework for Autonomous Vehicles
}
\author{Haoyang Fan\(^{1}\), Zhongpu Xia\(^{2}\), Changchun Liu\(^{2}\), Yaqin Chen\(^{2}\) and Qi Kong \(^{1,*}\)
\thanks{\(^{1}\) Haoyang Fan and Qi Kong are with Baidu USA LLC.}%
\thanks{\(^{2}\) Zhongpu Xia, Changchun Liu and Yaqin Chen are with Baidu Inc.}%
 \thanks{\(^{*}\) Corresponding author for this manuscript}%
}

\maketitle
\thispagestyle{empty}
\pagestyle{empty}


\begin{abstract}
 Many autonomous driving motion planners generate trajectories by optimizing a reward/cost functional.
 Designing and tuning a high-performance reward/cost functional for Level-4 autonomous driving vehicles with exposure to different driving conditions is challenging.
 Traditionally, reward/cost functional tuning involves substantial human effort and time spent on both simulations and road tests.
As the scenario becomes more complicated, tuning to improve the motion planner performance becomes increasingly difficult.
To systematically solve this issue, we develop a data-driven auto-tuning framework based on the Apollo autonomous driving framework.
The framework includes a novel rank-based conditional inverse reinforcement learning algorithm, an offline training strategy and an automatic method of collecting and labeling data. Our auto-tuning framework has the following advantages that make it suitable for tuning an autonomous driving motion planner.
First, compared to that of most inverse reinforcement learning algorithms, our algorithm training is efficient and capable of being applied to different scenarios.
Second, the offline training strategy offers a safe way to  adjust the parameters before public road testing.
Third, the expert driving data and information about the surrounding environment are collected and automatically labeled, which considerably reduces the manual effort.
Finally, the motion planner tuned by the framework is examined via both simulation and public road testing and is shown to achieve good performance.
\end{abstract}

\section{Introduction}
\subsection{Background}
A motion planner for autonomous driving aims to generate a safe and comfortable trajectory that leads an autonomous vehicle to the desired destination, which is a challenging and attractive task for both academia and industry.
Typically, autonomous driving motion planners need to understand the environment before sending trajectories to the vehicle control module. The step of understanding the environment is usually achieved through extracting features that capture the ego vehicle state, interactions with obstacles, traffic regulation constraints, etc. These features together form the state of the ego car. Then, the motion planner establishes a map from the state space of the current environment to the vehicle moving trajectory space.
\subsection{Related Work}
Typically, two major approaches are used to develop such a map: learning via demonstration (imitation learning) or through optimizing the current reward/cost functional.
\paragraph{Imitation learning}
In an imitation learning system, the state-to-action mapping is directly learned from expert demonstration.
A wide range of applications (e.g., \cite{bojarski2016end} and \cite{pomerleau1991efficient}) have been proposed and demonstrated to be effective due to the straightforward supervised learning framework.
However, direct application of imitation learning to a complicated robotic system such as autonomous driving faces some difficulties.
First, imitation learning lacks a generic understanding of the environment, works for only limited or simple scenarios and requires a large amount of collected information. The quantity, quality, and coverage of data are all critical to imitation learning.
The behavior of autonomous driving is hard to predict when applied to new scenarios.
Second, imitation learning has to give special attention to the covariate shifting issue. The environment may change dramatically as time passes. 
Modifications such as \cite{ross2010efficient}, \cite{ross2011reduction} and \cite{sun2017fast} have been proposed to solve this issue. However these methods usually required more collected demonstration data from an expert. The related data collection process is usually not efficient for large-scale problems. 
Additionally, in some applications, such as autonomous driving, scenarios or states are also hard to reproduce, since such applications usually involve considerable interaction with surrounding obstacles as well as constraints.
An imitation learning approach is difficult to maintain from a system perspective when we consider Level-4 autonomous driving systems.
An end-to-end imitation learning framework system is difficult to consider, especially when improper behavior occurs. 
Other problems, such as multimodal distributions, may also slow the training process. For example, when training data include human expert demonstration of nudging an obstacle either from left or right, the expert might pick up either one as a driving trajectory. When both sides of the trajectory exist in the training dataset, a multimodal distribution loss function is necessary but will slow the training process.

\paragraph{Optimizing through a reward functional}
Generating driving actions through maximizing a reward functional is more generic.
A wide range of traditional motion planning approaches, such as \cite{montemerlo2008junior}, \cite{urmson2008autonomous}, \cite{werling2010optimal} and \cite{ziegler2014trajectory}, derive their policies with a prespecified reward/cost functional. These approaches either discretize the space into a lattice and apply search methods such as dynamic programming or directly optimize through numeric optimization. The reward/cost functionals are typically provided by an expert or learned from data via inverse reinforcement learning.

\paragraph{Inverse reinforcement learning}
Inverse reinforcement learning (IRL) learns the reward functional by comparing the expert demonstration with generated trajectories or policies that optimize the reward functional (see \cite{russell1998learning}, \cite{ng2000algorithms}). In \cite{abbeel2004apprenticeship} the authors proposed reward function learning via feature expectation matching, while \cite{ratliff2006maximum} extended to a generalized maximum margin optimization problem. However, optimization through feature expectation matching is ambiguous; it requires optimized policies lying in the policy subspace that match the demonstrated behavior even when the behavior is not optimal. To solve this issue, \cite{ziebart2008maximum} proposed an IRL framework based on maximizing the cross entropy loss and demonstrated the performance by solving a routing problem. However, most IRL methods are computationally expensive since reinforcement learning or sampling is required to generate policies in every iteration of reward functional updating.

Many of the current IRL approaches perform well on a specific task with a fair mount of expert data and training time. However, some challenging aspects remain when applying these learning-based methods to autonomous driving motion planning problems.
First, autonomous driving systems require public road safety.
Public road test safety is important during both training and testing.
Many learning-based methods require substantial online training time to collect feedback from real-world driving, which may risk road safety.
Second, autonomous driving data are hard to reproduce. Expert driving data from different scenarios are easy to collect but are extremely difficult to reproduce in simulation since the ego car requires interaction with the surrounding environment. 
Finally, the motion planner for autonomous driving must not only meet the vehicle dynamic requirements but also follow traffic regulations at all times. How to systematically combine these constraints for reinforcement learning is not straightforward.
All these characteristics make data-driven motion planning a challenging task for the autonomous driving motion planner.

\subsection{Tuning Motion Planner for Autonomous Driving}
\begin{figure}[htbp]
\begin{center}
\includegraphics[width =  0.9\linewidth]{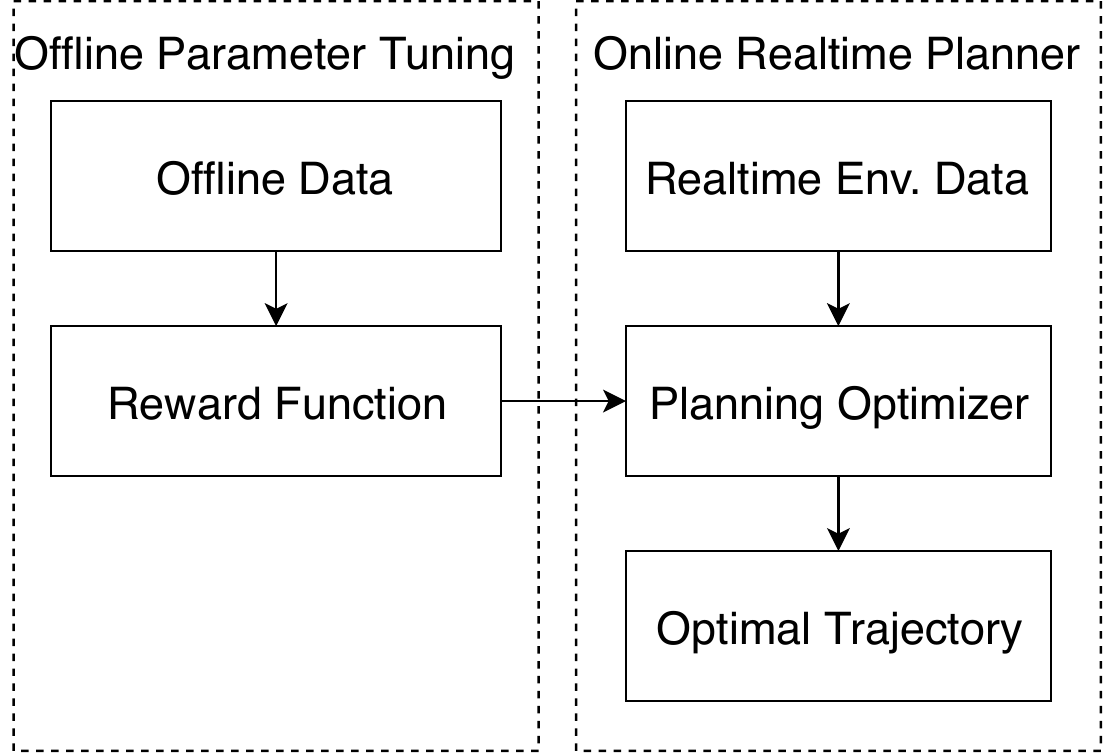}
\caption{Data-driven autonomous driving motion planner on the Apollo platform.}
\label{fig:autotuningmp}
\end{center}
\end{figure}

To scale up motion planner scenario coverage and improve case performance, we build an auto-tuning system that includes both online trajectory optimization and offline parameter tuning, as shown in Fig.~\ref{fig:autotuningmp}.

In the online module, we focus on yielding an optimal trajectory given a reward functional under constraints. Our motion planner module is not tied to a specific approach. One can use a different motion planner, such as sampling-based optimization, dynamic programming or reinforcement learning, to generate the trajectories. The performance of these motion planners is be evaluated with the metrics that quantify both optimality and robustness. Typically, the optimality of the online part can be measured by the difference in the reward functional values of the optimal trajectory and  generated trajectory, and the robustness can be measured by the variance in the generated trajectory behavior given specific scenarios. Simulations and road tests provide the final assessment of motion planner performance.

\begin{figure}[htbp]
\begin{center}
\includegraphics[width = \linewidth]{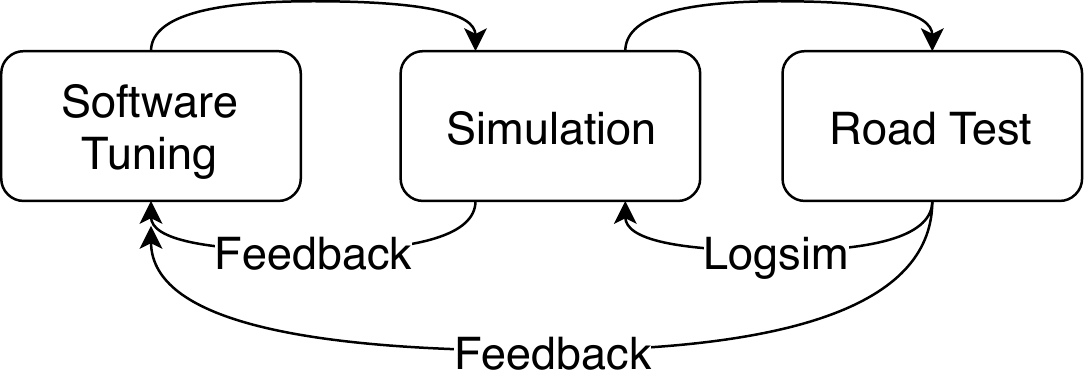}
\caption{Algorithm tuning loop for the motion planner in the Apollo autonomous driving platform.}
\label{fig:autotuningsim}
\end{center}
\end{figure}

For the offline tuning module, we focus on providing a reward/cost functional that can adapt to different driving scenarios.
A motion planning reward/cost functional contains features that describe the smoothness and interactions with the surrounding environment.  Typically, the reward/cost functional  can be tuned via both simulations and road testing. As shown in Fig.~\ref{fig:autotuningsim}, testing for a set of parameters requires both simulation and on-road testing. However, feedback cycles are the most time-consuming component since thousands or more driving scenarios are needed before drawing a conclusion from only one set of parameters. 

The aforementioned autonomous driving scenarios vary across many different driving conditions, including city, urban, highway and crowded regions.
Tuning a reward functional to adapt to these differences is difficult.
Traditionally, one starts with tuning simple scenarios and then extending to complicated ones.
When the current reward functional does not perform well in a new scenario, additional fine tuning and parameter extension may become necessary but will slow the process.  Furthermore, the features that are used to build the reward functional may be collinear, which may also impact the stability of the tuned motion planner. Thus, parameter tuning by experts becomes increasingly intractable, and a framework that can systematically solve this issue is urgently needed.

In this paper, we introduce a novel rank-based conditional IRL framework specifically targeting autonomous driving motion planner reward/cost functional tuning. The training process is offline and suitable for both large-scale testing and handling long-tail corner cases. The rest of this paper is organized as follows: Section \ref{sec:method} introduces the rank-based conditional IRL from the perspective of a Markov decision process (MDP). Section \ref{sec:framework} introduces the architecture of the auto-tuning framework based on the Apollo autonomous driving platform. In section \ref{sec:case_study}, we provide an example of speed profile tuning and compare the results with the idea of a general adversarial network (GAN). Section \ref{sec:conclusion} summarizes the paper and the results.

\section{Rank-based Conditional Inverse Reinforcement Learning (RC-IRL)}\label{sec:method}

\subsection{Preliminaries}
An MDP is defined by a set of states $\mathcal S$, transition actions $\mathcal{A}$ and transition probabilities $T = {P(s_{t+1} | s_t, a)}$. In most reinforcement learning frameworks, a reward functional $r \in \mathcal R$ is defined as a mapping $\mathcal S \to  \mathbb R$. 
A policy $\pi \in \Pi$ is defined as a map from a state to action $\pi(a_t | s_t)$ distribution, where $\Pi$ is the set of stationary policies. Reinforcement learning aims to find the policy that optimizes the accumulated reward function: 
\begin{equation} \hat \pi  = \arg \max_{\pi} \mathbb E_{s_0 \sim D}(V^{\pi}(s_0)),\end{equation}
where $s_0 \sim D$ is the initial state that follows predefined distribution $D$. $\mathbb E_{s_0 \sim D}(V^{\pi}(s_0))$ is defined as $\mathbb E \sum_{t = 0}^{\infty} \gamma_t r(s_t, a_t)$ with $a_t \sim \pi( \cdot | s_t), s_{t+1}  \sim P(\cdot | s_t, a_t)$, where $\gamma_t$ is a time discount factor. Finding a policy that optimizes the accumulative reward functional through the above method can be computationally expensive since sampling is required in every iteration. In reinforcement learning, the reward functional is usually provided by an expert or through IRL with expert demonstrations.

Define the expert policy as $\pi_{E}$. The idea of IRL is to find the reward functional such that the expected value function is best for the expert demonstration. The basic idea can be defined as follows:

\begin{equation}\label{eqn:irl}
\max_{r \in \mathcal R} (\mathbb E V^{\pi_E}(s_0) - \max_{\pi \in \Pi} \mathbb E V^{\hat \pi_r}(s_0) )
\end{equation}
where $\hat \pi_r$ is the estimated optimal policy generated by reinforcement learning under reward $r$. Some approaches (e.g., \cite{ratliff2006maximum}) refine the above formula, but the training process remains computationally expensive.

\subsection{RC-IRL}
Our idea for learning the reward functional includes two key parts: conditional comparison and rank-based learning.
\paragraph{Conditional comparison}
The expectation of value function can be rewritten as an integration of the initial state distribution:
\[\mathbb E_{s_0 \sim D}(V^{\pi}(s_0)) = \int_{D}V^{\pi}(s_0) P(s_0) d s_0  \]
These initial states may vary significantly for autonomous driving. Under these conditions, comparing the average behavior over the initial state distribution may not be efficient.
Thus, instead of comparing the expectation of value functions of the expert demonstration and optimal policy defined in Eq. \ref{eqn:irl}, we compare the value functions state by state. We use 
\begin{equation} \label{eqn:condition}\mathbb E_{s_0 \sim D} L(V^{\pi_E}(s_0) - V^{\pi_r}(s_0)) \end{equation}
to measure the performance of a policy $\pi_r$ under initial state $s_0$ given reward function $r(s, a)$. A loss function that is conditional on states can significantly reduce the background variance.

\begin{figure*}[htbp]
\begin{center}
\includegraphics[width = 0.245\linewidth]{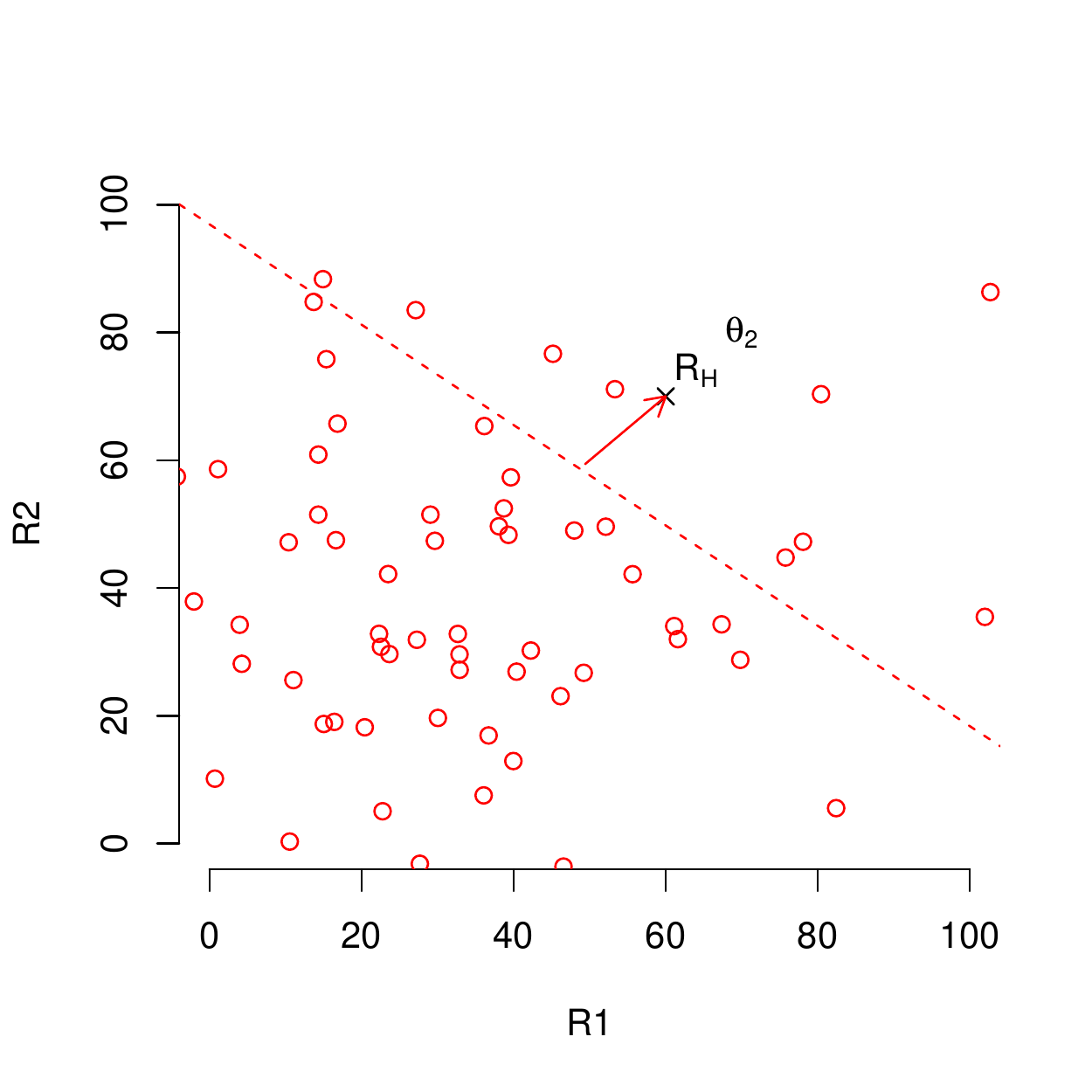}
\includegraphics[width = 0.245\linewidth]{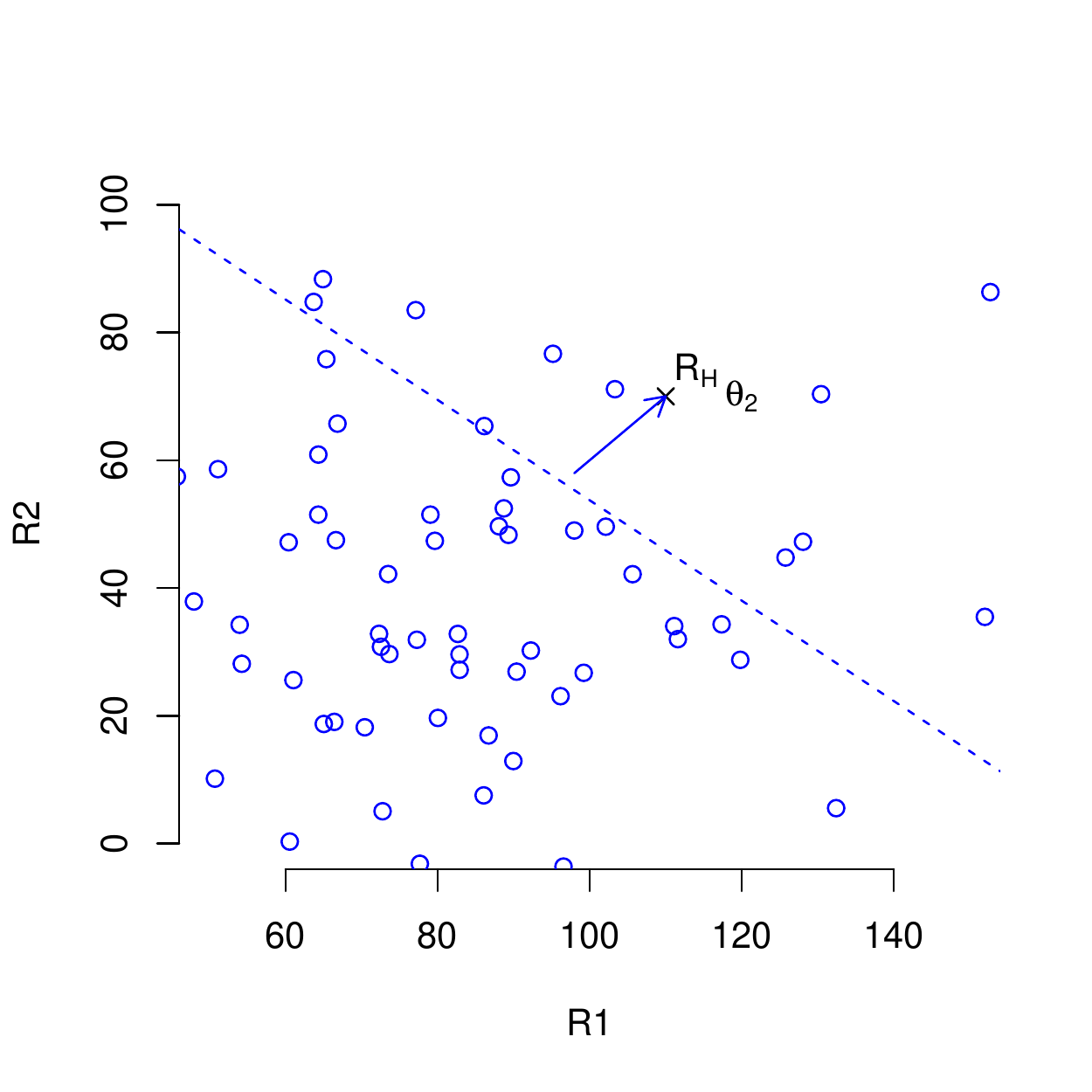}
\includegraphics[width = 0.48\linewidth]{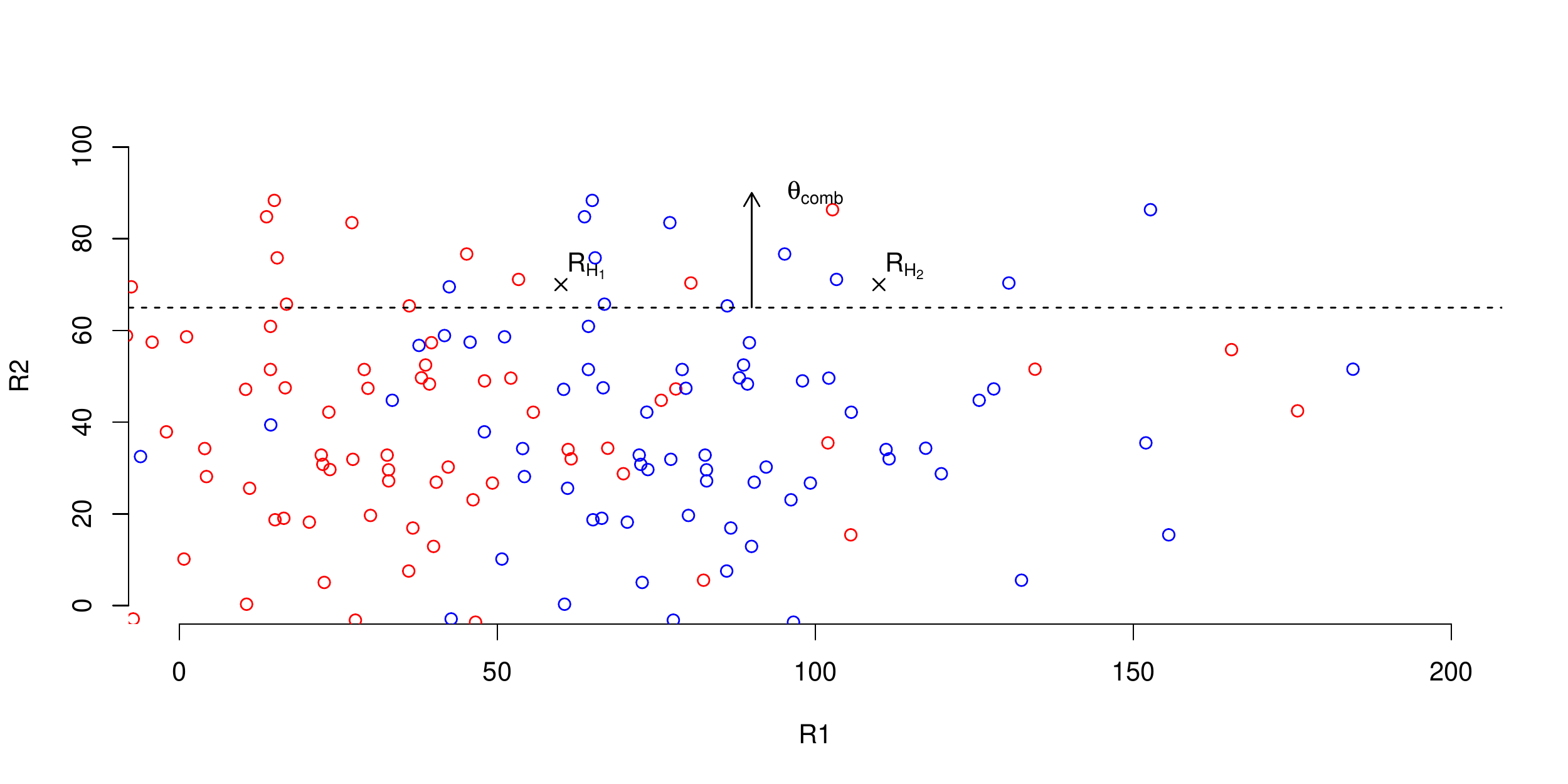}
\caption{Background shifting example. $\mathbf R_1$ and $\mathbf R_2$ represent two features of the reward function. The output reward function is a linear combination of $\mathbf R_1$ and $\mathbf R_2$. Thus, the reward function can be seen as the direction that maximizes the margin between pseudo demonstration $\mathbf R_H(\theta)$ and randomly generated samples. The circle points in the top-left of the figures are 100 samples randomly generated from a Cauchy distribution. The top-right figure shifts both the random samples and the pseudo demonstration point $\mathbf R_H$ by a fixed amount. The two red arrows represent the optimal direction for the top two frames. If two frames are combined, then the optimal direction shifts to the black one, as shown in the bottom figure. However, the direction trained with the combined frame is not optimal in either of the top frames.}
\label{fig:backgroundshift}
\end{center}
\end{figure*}

\paragraph{Rank-based learning}
To accelerate the training process and extend the coverage of corner cases, we sample random policies and compare against the expert demonstration instead of generating the optimal policy first, as in policy gradient. In detail, under each initial state $s_0$, which we define as a scenario, a set of random policies $\pi_{i}, i = 1, 2, ..., N$ is sampled. We compare the expert behavior over those policies given the current scenarios. These random policies can also be rephrased as random trajectories or random trajectory distributions for autonomous driving motion planning.
Our assumption is that the human demonstrations rank near the top of the distribution of policies conditional on initial state $s_0 \in D$ on average. Thus, the following expected conditional difference can be used as a loss function to optimize the reward functional:
\begin{equation} \mathbb E_{s_0 \sim D}  \sum_{i = 1, ..., N} L(V^{\pi_E}(s_0) - V^{\pi_i}(s_0)).\end{equation}
The above form can easily  adapt to refinement and improvement of the loss function and training, for example,  \cite{ratliff2006maximum} and \cite{ziebart2008maximum}. Additionally, since we generate random policies for comparison with the expert demonstration, the tuned reward functional can easily learn useful information from corner cases. Difficult scenarios can also be generated to train and test the robustness of the reward functional.
\paragraph{Background shifting problem}
Our idea of a conditional comparison instead of expectation comparison is based on the complexity of the autonomous driving motion planning problem. Since our motion planner has to address different driving scenarios, such as highway, local, and heavy traffic, substantial differences in the behavior metrics may exist. These background differences may impact the tuned reward functional significantly. We illustrate the issue of background shifting with an example shown Fig.~\ref{fig:backgroundshift}, where two frames with different $s_0 \in D$ are sampled. In each frame, 100 randomly generated trajectories are sampled for comparison with human-demonstrated trajectories. Based on the idea of the maximum margin \cite{ratliff2006maximum}, the goal is to find the direction that clearly separates the demonstrated trajectory from randomly generated ones. However, even if the optimal reward function direction is the same under the two scenarios, it may not be ideal to train them together because the optimal direction may be impacted by overfitting the background shifting. Instead, the idea of conditioning on scenarios can be viewed as a pairwise comparison, which can remove the background differences.
\section{Auto-tuning Implementation in the Apollo Autonomous Driving System}\label{sec:framework}
\subsection{Framework}
\begin{figure}[htbp]
\begin{center}
\includegraphics[width = 0.9\linewidth]{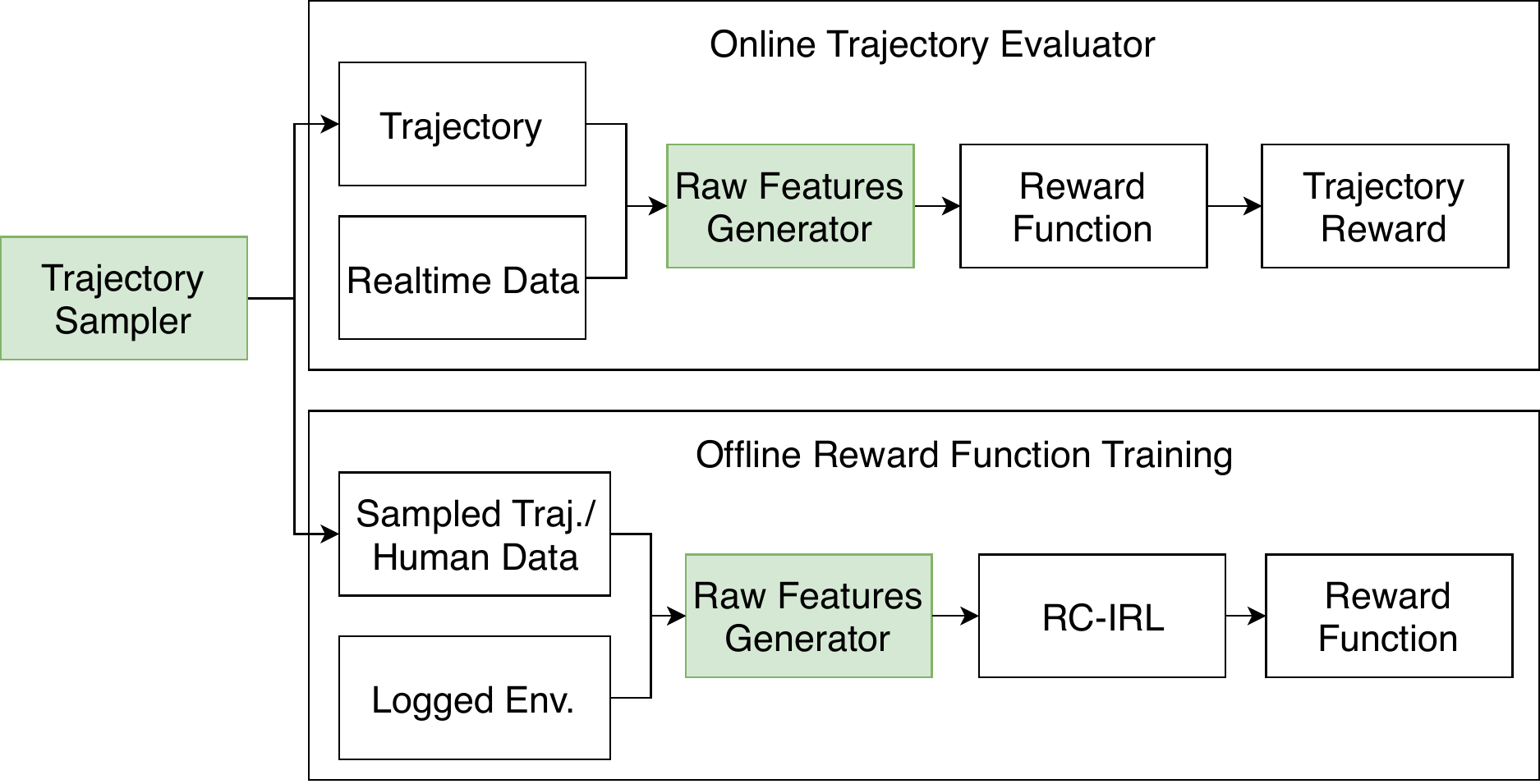}
\caption{Online and offline pipeline of the auto-tuning framework by module.}
\label{fig:pip}
\end{center}
\end{figure}

Auto-tuning in the Apollo autonomous driving system involves both online trajectory evaluation and offline parameter tuning, as shown in Fig.~\ref{fig:pip}. The two components share some common modules for the purpose of consistency. The raw feature generator takes input from the environment and evaluate sampled or human expert driver trajectories indiscriminately; the trajectory sampler uses the same strategy to generate candidate trajectories for both the offline and online modules. In the online evaluator, after the raw features are extracted from a trajectory, a reward/cost functional is applied to provide a score. The final output trajectory is selected by ranking all the scored trajectories or through dynamic programming, such as a search-based algorithm.

\subsection{Training the Value Functional With a SIAMESE Network} 
For our motion planner, we define a trajectory under MDP as $\xi = (a_{0}, s_{0}, ... a_{N}, s_{N}) \in \Xi$, where space $\Xi$ is the sampled trajectory space. A trajectory under initial state $s_0$ is evaluated by the value function:
\begin{equation}\label{eqn:value} V^{\xi}( s_0)  = \sum_{t = 0}^{N} \gamma_{t} R(a_t, s_t),\end{equation}
which is a linear combination of the rewards at various time points. The raw feature generator module provides a sequence of features based on the current state and action. We use
$f_{j}(a_t, s_t), j = 1, 2, ..., K$ to represent the feature given a current state and action. We choose reward function R as a function of all features with parameter $\theta \in \Omega$:

\begin{equation}\label{eqn:reward}R_{\theta} (a_t, s_t)= \tilde R(f_1, ..., f_K, \theta).\end{equation}
Typically, $\tilde R$ can be as simple as a linear combination of features or a neural network with features as the input. The latter can be treated as a feature-encoding procedure to further capture the intrinsic characteristics of state-to-action mapping. The value function is a rank or search objective for selecting best trajectories in the online module.

\begin{figure}[htbp]
\begin{center}
\includegraphics[width = \linewidth]{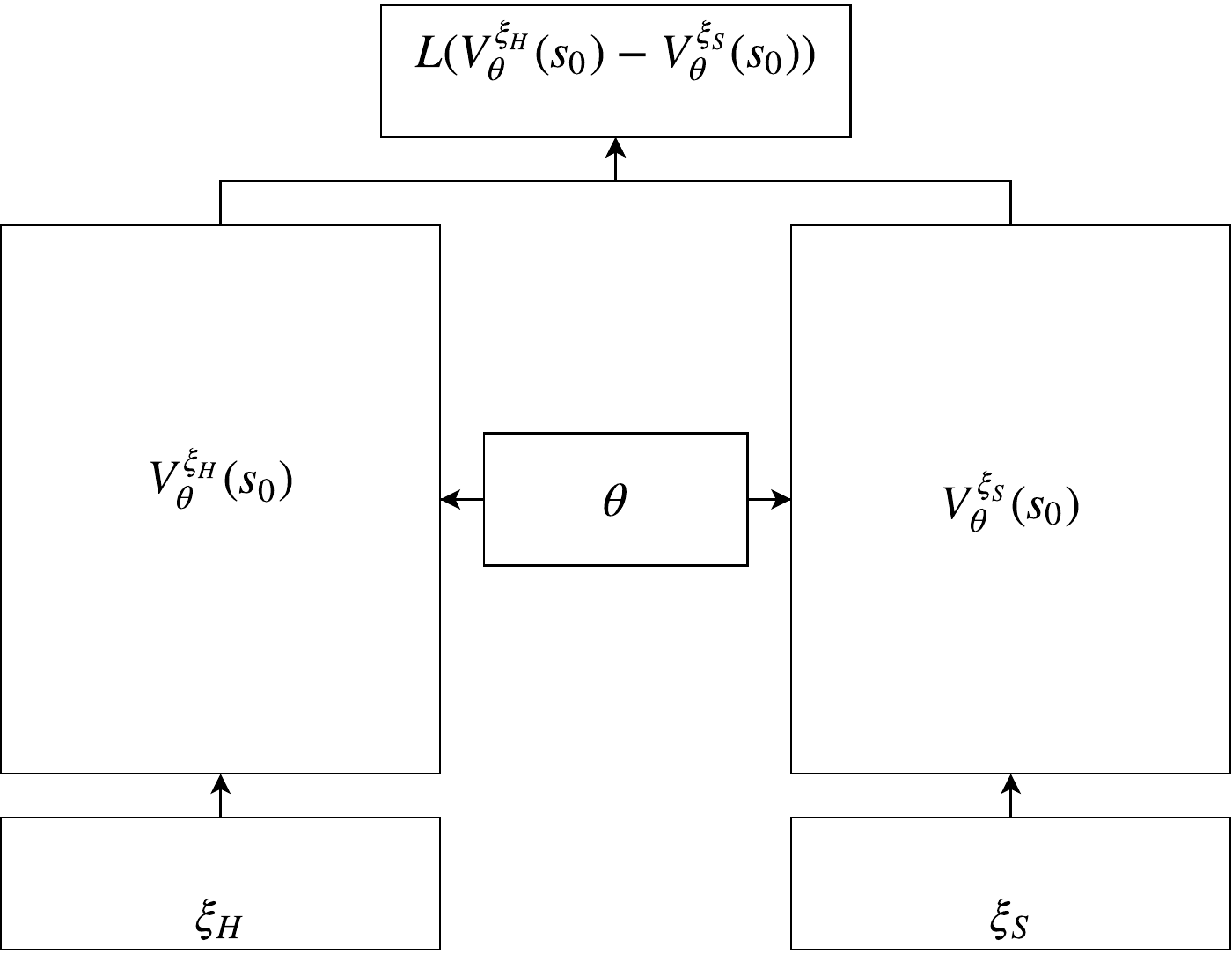}
\caption{Siamese network in RC-IRL. The value networks of both the human and the sampled trajectories share the same network parameter settings. The loss function evaluates the difference between the sampled data and the generated trajectory via the value network outputs.
}
\label{fig:siamese}
\vspace{-1em}
\end{center}
\end{figure}

We use the aforementioned RC-IRL method for the offline module.
 We use $\xi_{H}$ to represent the human expert demonstration trajectory and $\xi_S$ to represent the randomly generated sample trajectory in $\Xi$. The learning procedure has no differences compared to training the SIAMESE network \cite{chopra2005learning} since RC-IRL's loss function essentially compares the output of a value network. The structure of SIAMESE in RC-IRL is presented in Fig. ~\ref{fig:siamese}.

\section{Case studies}\label{sec:case_study}

In this section, we present an application of speed profile generation inside EM 
planner \cite{2018arXiv180708048F}. Given a path profile in a station-lateral coordinate system, obstacles and predicted moving trajectories are projected on the station-time graph if there are any interactions with the moving path of the ego car. Then, the goal is to generate a speed profile on the station-time graph that can safely avoid obstacles and maintain smooth driving.  The optimal speed profile is generated by optimizing the cost/reward functional, which captures the trajectory smoothness, distance to different obstacles, and path smoothness.

\subsection{Model Setting}
\begin{figure}[htbp]
\begin{center}
\includegraphics[width = \linewidth]{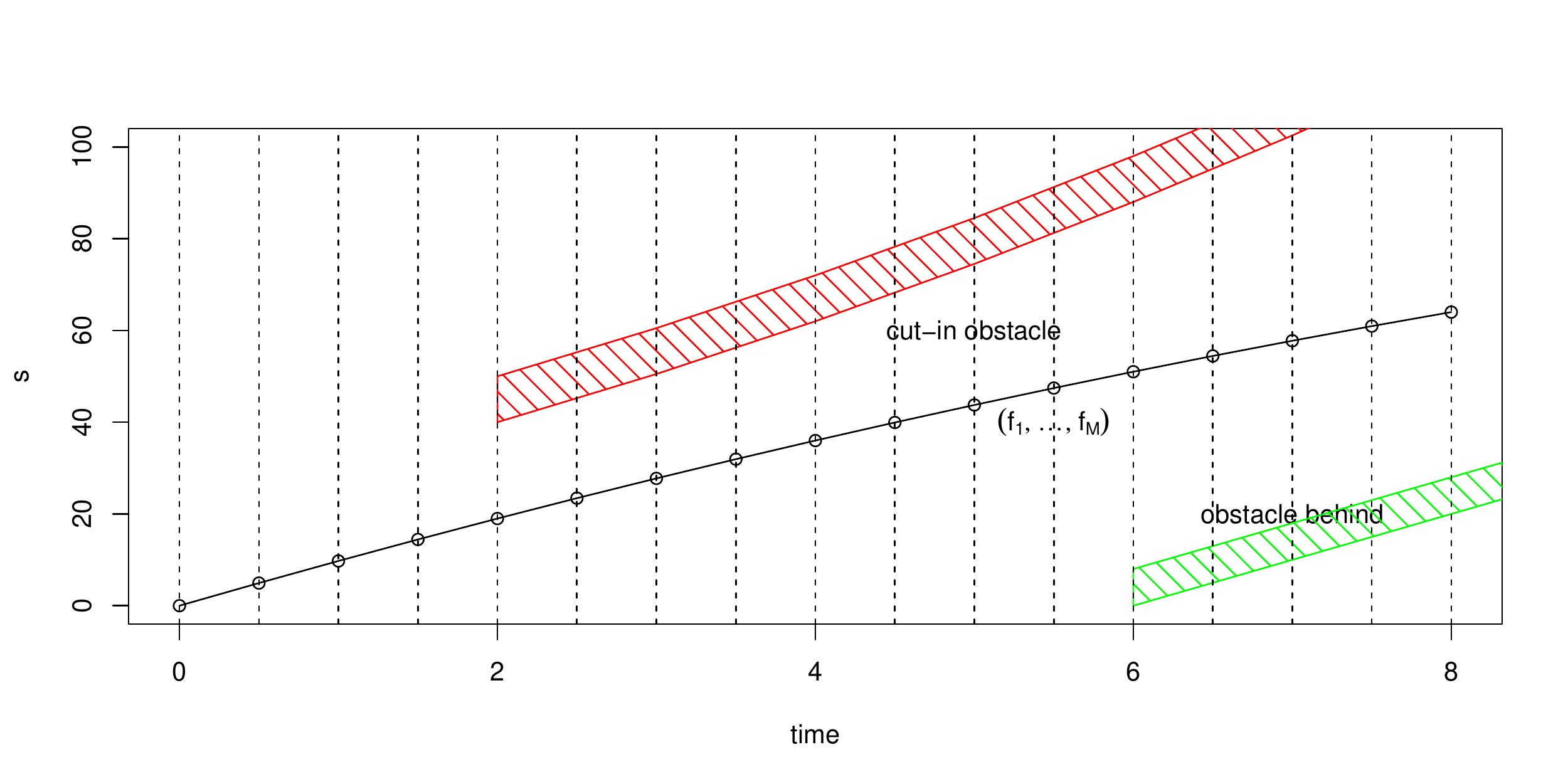}
\vspace{-2em}
\caption{Feature extraction on the station-time graph.}
\label{fig:st_graph}
\end{center}
\vspace{-2em}
\end{figure}

\begin{figure}[htbp]
\begin{center}
\includegraphics[width = 0.9\linewidth]{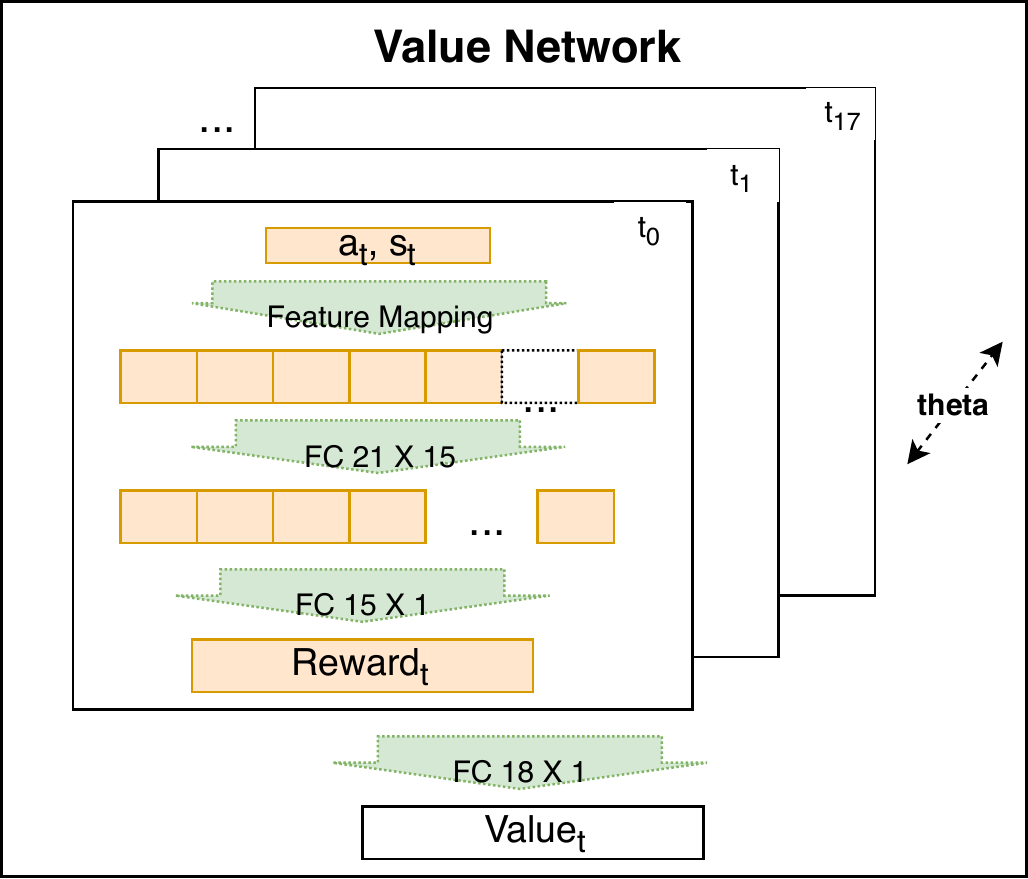}
\end{center}
\caption{The value network inside the siamese model is used to capture driving behavior based on encoded features. The network is a trainable linear combination of encoded rewards at different times $t = t_0, ..., t_{17}$. The weight of the encoded reward is a learnable time decay factor.
The encoded reward includes an input layer with 21 raw features and a hidden layer with 15 nodes to cover possible interactions. The parameters of the reward at different times share the same $\theta$ to maintain consistency.} \label{fig:value_network}
\end{figure}

As shown in Fig.~\ref{fig:st_graph}, the reward/cost functional is evaluated at fixed time points  $t_0, t_1, ..., t_N$ given a trajectory $\xi$. Denote $(s_t, a_t)$ as $\xi$'s trajectory point at the corresponding time. The definitions of the reward and value functions follow Eq. \ref{eqn:reward} and Eq. \ref{eqn:value}.  We provide a detailed description of the features in Table~\ref{tab:features}.
\begin{table}[htp]
\caption{Feature Description Given A Trajectory Point}\label{tab:features}
\begin{center}
\begin{tabular}{| l |p{0.65\linewidth}|}
\hline
\textbf{Feature} & \textbf{Description} \\
\hline
\textbf{l} & lateral coordinate w.r.t. lane center \\ \hline
\textbf{dl} & derivative of lateral coordinate \\ \hline
\textbf{ddl} & second-order derivative of lateral coordinate \\ \hline
\textbf{curvature} & curvature \\  \hline
\textbf{station} & station position of current car location \\ \hline
\textbf{time} & time \\ \hline
\textbf{velocity} & current vehicle velocity \\ \hline
\textbf{speed limit} & road speed limit at current trajectory point location \\ \hline
\textbf{acceleration} & acceleration at current trajectory point\\ \hline
\textbf{jerk} & jerk at current trajectory point\\ \hline
\textbf{collision dist.}& distance to closest obstacle\\ \hline
\textbf{follow obs.} & features with followed obstacle, including follow station distance and follow obstacle speed \\ \hline
\textbf{overtake obs.} & features with overtake obstacle, including station distance to overtake obstacle and overtake obstacle speed \\ \hline
\textbf{stop obs.} & stop line station distance\\ \hline
\textbf{virtual obs.} & virtual obstacle includes the destination of routing, similar as stop\\ \hline
\textbf{nudge obs.} & nudge obstacle lateral position difference and nudge obstacle speed\\ \hline
\end{tabular}
\end{center}
\label{default}
\end{table}
Note that $\gamma_t$ is not necessarily exponential decay since we expect that the model can learn trends from data.
The data used for training the auto-tuning model are collected from a human expert driving under various scenarios. The recorded data include the current surrounding environment, such as obstacle information, and information about the ego car, including position, velocity, acceleration, and jerk. This information is used later for extracting features by a predefined mapping $\mathbf F$ in the raw feature generator. The human expert trajectory and randomly generated sample trajectories are sent to a SIAMESE network in a pair-wise manner, as shown in Fig.~\ref{fig:siamese}. The architecture of the SIAMESE value network is shown in Fig.~\ref{fig:value_network}. Our SIAMESE network uses the leaky-RELU loss function:
\begin{equation}
L(y) = \begin{cases}
 L(y) = y, ~~ y \geq 0 \\
 L(y) = ay, ~~ y < 0
\end{cases} \end{equation}
with leaky rate $a = 0.05$.
Further, to validate the general performance of our idea in terms of conditional IRL, we consider a general adversarial network (GAN) for comparison. 
In GAN, each sampled trajectory is labeled as human driving $(1)$ or random trajectory $(0)$ and trained with cross entropy loss. We hope that the network can distinguish human driving from a randomly generated trajectory. Since the GAN cannot distinguish different scenarios, the procedure compares the average performance of human driving trajectories and that of randomly generated trajectories.

\subsection{Training Process}
Our training data were selected from approximately 1000+ hours of expert driving data by filtering frames with no obstacles or speed changes.
The data include 718 million frames of human driving records after filtering. The offline training module generated 2.8 billion corresponding queries of randomly sampled trajectories after filtering. Approximately 2 epochs of stochastic gradient descent were required to converge to an ideal reward functional. In our experiment, different training methods, such as ADAM or RMSProp, do not substantially affect the solution since the reward functional is simple.

\subsection{Experiments and Results} 
\begin{figure}[htbp]
\vspace{-1em}
\begin{center}
\includegraphics[width = 0.9\linewidth]{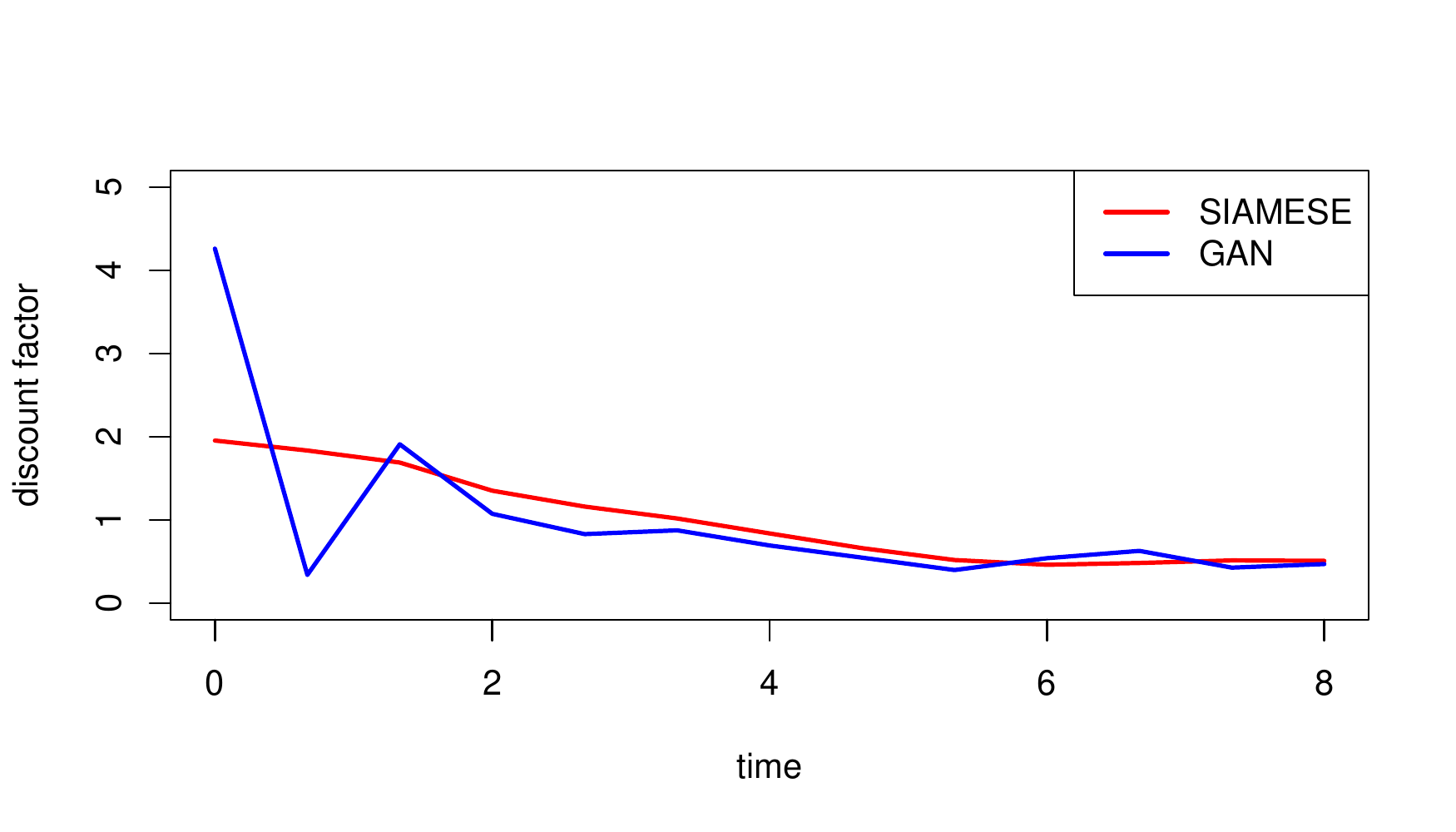}
\vspace{-1em}
\caption{The comparison of the trained time discount factor $\gamma_t, t \in [0, 8 s]$of SIAMESE and GAN under the auto-tuning framework. The time discount factor trained by the SIAMESE network performed well compared to that trained by GAN.}
\label{fig:comp}
\vspace{-2em}
\end{center}
\end{figure}

The training performance of the SIAMESE and GAN networks were evaluated by scenario-based simulation tests. The simulation test includes 3400+ cases that covers stopping, turning, changing lanes, yielding, overtaking and more complicated scenarios. Each case lasts approximately 2 minute and is measured by performance metrics. We list the key performance metrics relative to the motion planner and the results in Table~\ref{tab:comparison}. In the table, we list $P_{\text{collision free}}$ as the safety index of the motion planner and the probability of lateral and longitudinal acceleration and jerk constraints within range as an index of trajectory smoothness. In our simulation, SIAMESE with RC-IRL performs better than the reward functional trained by GAN.
 Fig.~\ref{fig:comp} shows the distributions of the trained discount factors of the two methods.  As shown, the training performance based on the SIAMESE network is better than that of the GAN network.
 As of July 25, 2018, our tuned reward functional in the Apollo platform has been tested on the road and has shown good performance, with 25000+ miles of driving on the road since April 2, 2018. 

\begin{table}[htbp]
\caption{Simulation Results of the Motion Planner}
\begin{center}
\begin{tabular}{ | p{0.5\linewidth} |  r | r |}\hline
\textbf{Metric} & SIAMESE & GAN \\ \hline
$\mathbf P_{\text{collision free}}$ & $$100\%$$ & $100\%$\\ \hline
$\mathbf P(V < V_{\text{limit}})$ &  $100\%$& $99.49\%$\\ \hline
$\mathbf P(|a_{\text{station}}| < 4.0) $ & $99.33\%$& $86.80\%$\\ \hline
$\mathbf P(|a_{\text{lateral}}| < 4.0)  $ & $99.97\%$ & $99.97\%$\\ \hline
$\mathbf P(|j_{\text{station}}| < 6.0)$ & $97.77\%$ & $71.20\%$\\ \hline
$\mathbf P(|j_{\text{lateral}}| < 6.0)$ &$82.77\%$ & $82.57\%$ \\ \hline
\end{tabular}
\end{center}
\label{tab:comparison}
\end{table}

\begin{figure}[htbp]
\begin{center}
\includegraphics[width = 0.9\linewidth]{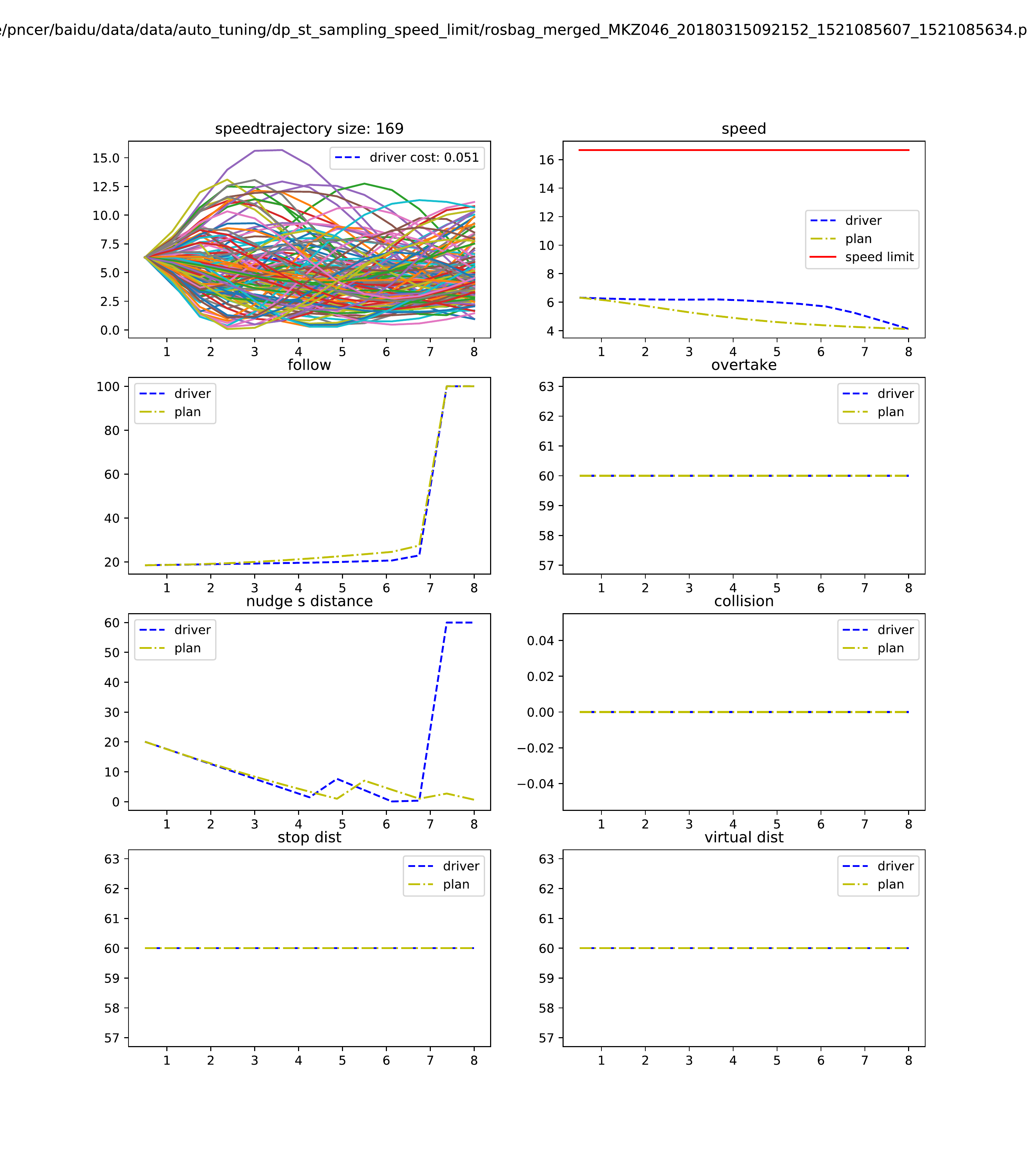}
\caption{One frame of the optimal trajectory with respect to the tuned reward functional compared to the human driving trajectory. The top-left figure represents the randomly generated sample distribution of the speed versus time graph. The top-right figure compares the speed profiles of the human driver and the generated speed profile over time. The remaining figures describe the distance to different types of obstacles for the human driving data and optimal speed profile w.r.t. the tuned reward functional.}
\label{fig:example}
\vspace{-1em}
\end{center}
\end{figure}

For additional illustration, we extract one frame to examine the performance of the learned reward function in Fig.~\ref{fig:example}.
\section{Conclusion} \label{sec:conclusion}
In this paper, we proposed an auto-tuning framework for a Level-4 autonomous driving motion planner based on the Apollo autonomous driving platform \url{https://github.com/ApolloAuto/apollo}. The proposed method includes automatic human driving data labeling, a scalable tuning framework and a novel rank-based conditional IRL. Compared to existing tuning approaches, our data-driven approach efficiently makes use of demonstrated data and easily adapts to different driving scenarios. The method is suitable for both large-scale training and handling long-tail corner cases. Further research can be conducted to refine the design of the value network, loss function, and feature extraction.
\section*{Acknowledgement}
We would like to thank Apollo Community for their useful suggestions and contributions.
\bibliographystyle{IEEEtran}
\bibliography{IEEEabrv,reference}
\end{document}